\documentclass[11pt,a4paper]{article}
\usepackage[hyperref]{acl2017}
\usepackage{times}
\usepackage{latexsym}
\usepackage{url}
\usepackage{graphicx}
\usepackage{amsfonts,amssymb}
\usepackage{amsmath}
\usepackage{colortbl}
\usepackage{color}
\definecolor{green}{rgb}{.2,.8,.1}
\definecolor{red}{rgb}{.9,.1,.1}
\definecolor{blue}{rgb}{.1,.1,.9}
\newcommand{\tabincell}[2]{\begin{tabular}{@{}#1@{}}#2\end{tabular}}
\makeatletter
\def\hlinewd#1{%
  \noalign{\ifnum0=`}\fi\hrule \@height #1 \futurelet
   \reserved@a\@xhline}
\makeatother
\aclfinalcopy 

\title{Joint Extraction of Entities and Relations \\Based on a Novel Tagging Scheme}
\author{Suncong Zheng, Feng Wang, Hongyun Bao,
Yuexing Hao,Peng Zhou,  Bo Xu
\\Institute of Automation, Chinese Academy of Sciences, 100190, Beijing, P.R. China
\\\{suncong.zheng, feng.wang,hongyun.bao, haoyuexing2014,  peng.zhou,xubo\}@ia.ac.cn}
\date{}

\begin{document}
\maketitle
\begin{abstract}
Joint extraction of entities and relations is an important task in information extraction.
To tackle this problem, we firstly propose a novel tagging scheme
that can convert the joint extraction task to a tagging problem.
Then, based on our tagging scheme, we study different end-to-end models to extract entities and their relations directly,
without identifying entities and relations separately.
We conduct experiments on a public dataset  produced by distant supervision method and
the experimental results show that the tagging based methods are
better than most of the existing pipelined and joint learning methods.
What's more, the end-to-end model proposed in this paper, achieves the best results on the public dataset.
\end{abstract}

\section{Introduction}
\label{sect:introduction}
\begin{figure}[h]
 \begin{center}
\includegraphics[width=8cm]{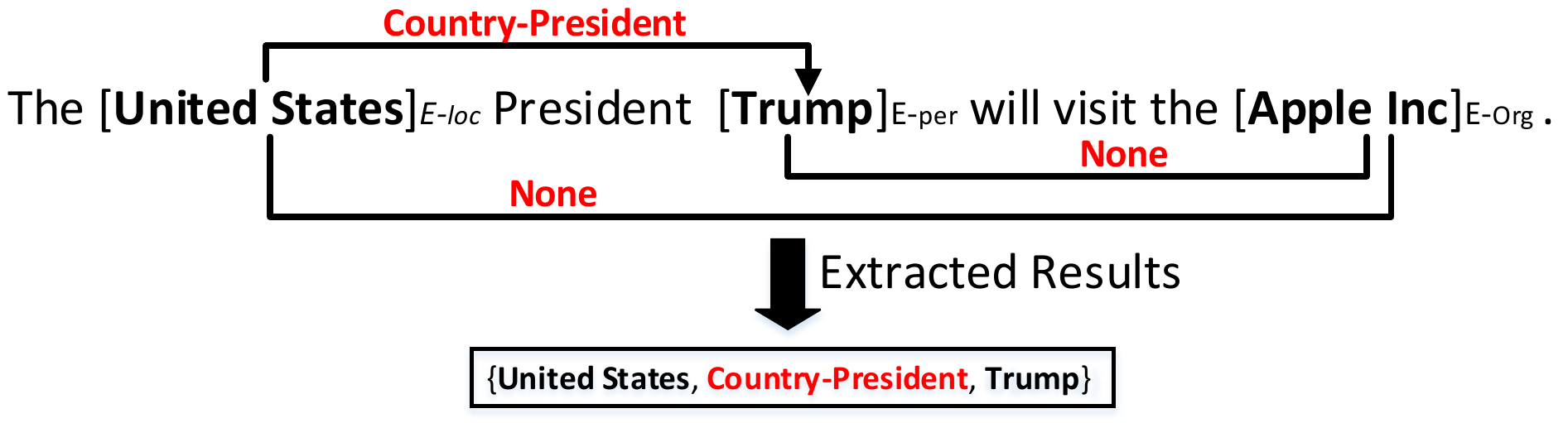}
\caption{\label{exp}A standard example sentence for the task. ``Country-President'' is a relation in the predefined relation set.}
 \end{center}
\end{figure}

Joint extraction of entities and relations is to detect entity mentions and
recognize their semantic relations simultaneously from unstructured text, as Figure \ref{exp} shows.
Different from open information extraction (Open IE) \cite{banko2007open} whose relation words are extracted from the given sentence,
in this task, relation words are extracted from a predefined relation set which may not appear in the given sentence.
It is an important issue in knowledge extraction and automatic construction of knowledge base.

Traditional methods handle this task in a
pipelined manner, i.e., extracting the entities \cite{nadeau2007survey}
first and then recognizing their relations \cite{fea2010utd}.
This separated framework makes the task easy to deal with, and each component can be more flexible.
But it neglects the relevance between these two sub-tasks and each subtask is an independent model.
The results of entity recognition may affect the performance of relation classification and lead to erroneous delivery \cite{li2014}.

Different from the pipelined methods,
joint learning framework is to extract entities together with relations using a single model.
It can effectively integrate the information of entities and relations, and
it has been shown to achieve better results in this task.
However, most existing joint methods are feature-based structured systems \cite{li2014,miwa2014modeling,yu2010colling,ren2016cotype}.
They need complicated feature engineering and heavily rely on the other NLP toolkits,
which might also lead to error propagation.
In order to reduce the manual work in feature extraction, recently,
\cite{miwa2016end} presents a neural network-based method for the end-to-end entities and relations extraction.
Although the joint models can represent both entities and relations with shared parameters in a single model,
they also extract the entities and relations separately and produce redundant information.
For instance, the sentence in Figure \ref{exp} contains three entities: ``United~States'', ``Trump'' and ``Apple~Inc''.
But only ``United~States'' and ``Trump'' hold a fix relation ``Country-President''.
Entity ``Apple~Inc'' has no obvious relationship with the other entities in this sentence.
Hence, the extracted result from this sentence is \{$\textbf{United~States}_{e1}$, $\textbf{Country-President}_{r}$, $\textbf{Trump}_{e2}$\}, which called triplet here.

In this paper, we focus on the extraction of triplets that are composed of two entities and one relation between these two entities.
Therefore, we can model the triplets directly, rather than extracting the entities and relations separately.
Based on the motivations, we propose a tagging scheme accompanied with the end-to-end model to settle this problem.
We design a kind of novel tags which contain the information of entities and the relationships they hold.
Based on this tagging scheme, the joint extraction of entities and relations can be transformed into a tagging problem.
In this way, we can also easily use neural networks to model the task  without complicated feature engineering.

Recently, end-to-end models based on LSTM \cite{lstm97} have been successfully applied to various tagging tasks:
Named Entity Recognition \cite{lstm-crf}, CCG Supertagging \cite{supertagging}, Chunking \cite{chunk} et al.
LSTM is capable of learning long-term dependencies, which is beneficial to sequence modeling tasks.
Therefore, based on our tagging scheme,
we investigate different kinds of LSTM-based end-to-end models to jointly extract the entities and relations.
We also modify the decoding method by adding a biased loss to make it more suitable for our special tags.

The method we proposed is a supervised learning algorithm.
In reality, however, the process of manually labeling a training set
with a large number of entity and relation is too expensive and error-prone.
Therefore, we conduct experiments on a public dataset\footnote{https://github.com/shanzhenren/CoType}
which is produced by distant supervision method \cite{ren2016cotype} to validate our approach.
The experimental results show that our tagging scheme is effective in this task.
In addition, our end-to-end model can achieve the best results on the public dataset.

The major contributions of this paper are:
(1) A novel tagging scheme is proposed to jointly extract entities and relations, which can easily transform the extraction problem into a tagging task. 
(2)  Based on our tagging scheme, we study different kinds of end-to-end models to settle the problem.
   The tagging-based methods are better than most of the existing pipelined and joint learning methods.
(3) Furthermore, we also develop an end-to-end model with biased loss function to suit for the novel tags.
    It can enhance the association between related entities.

\section{Related Works}
\label{sect:relatedwork}
Entities and relations extraction is an important step to construct a knowledge base, which can be benefit for many NLP tasks.
Two main frameworks have been widely used to solve the problem of extracting
entity and their relationships. One is the pipelined method and the other is the joint learning method.

The pipelined method treats this task as two separated tasks, i.e., named entity recognition (NER) \cite{nadeau2007survey}
and relation classification (RC) \cite{fea2010utd}.
Classical NER models are linear statistical models, such as
Hidden Markov Models (HMM) and Conditional Random Fields (CRF) \cite{passos2014lexicon, Gang2015Joint}.
Recently, several neural network architectures \cite{chiu2015named,huang2015bidirectional,lstm-crf} have been successfully applied to NER, which is regarded as a sequential token tagging task.
Existing methods for relation classification can also be divided into
handcrafted feature based methods \cite{fea2010utd,fea2004combining} and
neural network based methods \cite{dlpcnn2015semantic,zsc,cnn2014relation,lstm2015classifying,CR}.

While joint models extract entities and relations using a single model.
Most of the joint methods are feature-based structured systems
\cite{ren2016cotype,yang2013joint,singh2013joint,miwa2014modeling,li2014}.
Recently, \cite{miwa2016end} uses a LSTM-based model to extract entities and relations, which can reduce the manual work.

Different from the above methods, the method proposed in this paper is based on a special tagging manner,
so that we can easily use end-to-end model to extract results without NER and RC.
end-to-end method is to map the input sentence into meaningful vectors and then back to produce a sequence.
 It is widely used in machine translation \cite{13end2end,sutskever2014sequence}  and sequence tagging tasks \cite{lstm-crf,supertagging}.
Most methods apply bidirectional LSTM to encode the input sentences,
but the decoding methods are always different. For examples,
\cite{lstm-crf} use a CRF layers to decode the tag sequence, while \cite{supertagging,katiyarinvestigating} apply LSTM layer to produce the tag sequence.
\begin{figure*}
 \begin{center}
\includegraphics[width=15cm]{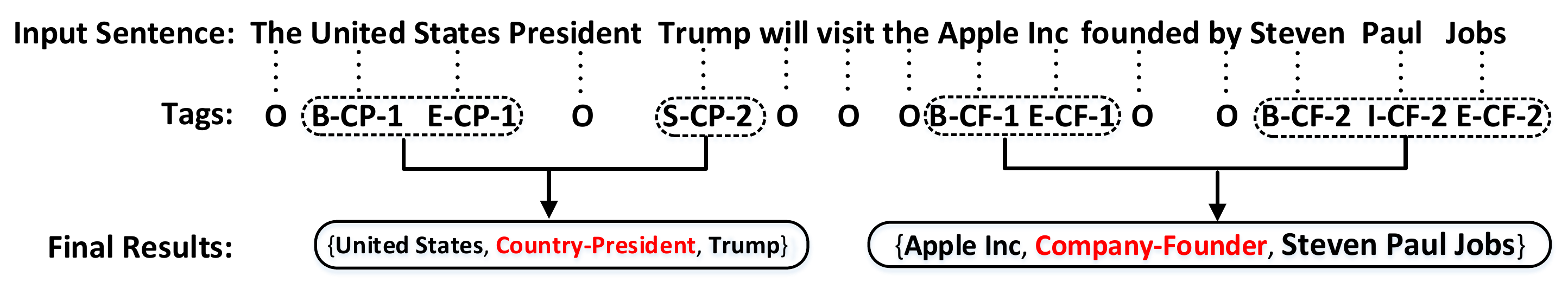}
\caption{\label{exp2}Gold standard annotation for an example sentence based on our tagging scheme, where ``CP'' is short for ``Country-President'' and ``CF'' is short for ``Company-Founder''.}
 \end{center}
\end{figure*}
\section{Method}
We propose a novel tagging scheme and an end-to-end model with biased objective function
to jointly extract entities and their relations.
In this section, we firstly introduce how to change the extraction problem to a tagging problem based on our tagging method.
Then we detail the model we used to extract results.
\subsection{The Tagging Scheme}
Figure \ref{exp2} is an example of how the results are tagged.
Each word is assigned a label that contributes to extract the results.
Tag ``O'' represents the ``Other'' tag, which means that the corresponding word is independent of the extracted results.
In addition to ``O'', the other tags consist of three parts:
the word position in the entity, the relation type, and the relation role.
We use the ``BIES'' (Begin, Inside, End,Single) signs to represent the position information of a word in the entity.
The relation type information is obtained from  a predefined set of relations and
the relation role information is represented by the numbers ``1'' and ``2''.
An extracted result is represented by a triplet: $(Entity_1,RelationType,Entity_2)$.
``1'' means that the word belongs to the first entity in the triplet,
while ``2'' belongs to second entity that behind the relation type.
Thus, the total number of tags is $N_t = 2*4*|R|+1$, where $|R|$ is the size of the predefined relation set.

Figure \ref{exp2} is an example illustrating our tagging method.
The input sentence contains two triplets:
\{\textbf{United~States, Country-President,  Trump}\} and \{\textbf{Apple Inc, Company-Founder, Steven Paul Jobs}\}, where
 ``Country-President'' and ``Company-Founder'' are the predefined relation types.
 The words ``United'',``States'',`` Trump'',``Apple'',``Inc'' ,``Steven'', ``Paul'' and ``Jobs'' are all related to the final extracted results. Thus they are tagged based on our special tags.
 For example, the word of ``United'' is the first word of entity ``United~States'' and
 is related to the relation ``Country-President'', so its tag is ``B-CP-1''.
The other entity `` Trump'', which is corresponding to ``United~States'', is labeled as ``S-CP-2''.
Besides, the other words irrelevant to the final result are labeled as ``O''.
 \subsection{From Tag Sequence To Extracted Results}
 From the tag sequence in Figure \ref{exp2},
 we know that `` Trump'' and ``United~States'' share the same relation type ``Country-President'',
 ``Apple Inc'' and ``Steven Paul Jobs'' share the same relation type ``Company-Founder''.
  We combine entities with the same relation type into a triplet to get the final result.
  Accordingly, `` Trump'' and ``United~States'' can be combined into a triplet whose relation type is ``Country-President''.
  Because, the relation role of `` Trump'' is ``2'' and ``United~States'' is ``1'', the final result is
\{\textbf{United~States, Country-President,  Trump}\}. The same applies to \{\textbf{Apple Inc, Company-Founder, Steven Paul Jobs}\}.

 Besides, if a sentence contains two or more triplets with the same relation type,
 we combine every two entities into a triplet based on the nearest principle.
 For example, if the relation type ``Country-President'' in Figure \ref{exp2} is ``Company-Founder'',
 then there will be four entities in the given sentence with the same relation type.
``United~States'' is closest to entity `` Trump'' and the ``Apple Inc'' is closest to ``Jobs'',
so the results will be \{\textbf{United~States, Company-Founder, Trump}\} and \{\textbf{Apple Inc, Company-Founder, Steven Paul Jobs}\}.

 In this paper, we only consider the situation where an entity belongs to a triplet,
 and we leave identification of overlapping relations for future work.

\subsection{The End-to-end Model}
\begin{figure*}[ht]
 \begin{center}
\includegraphics[width=15cm]{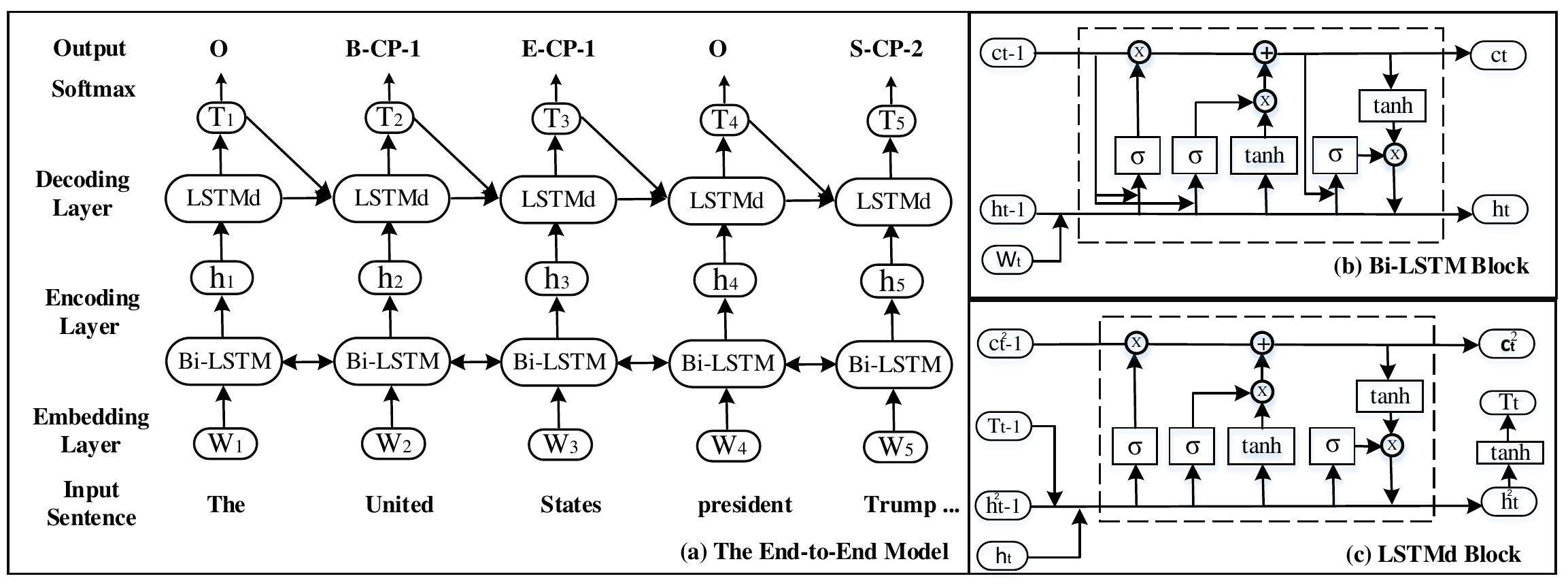}
\caption{\label{model}
An illustration of our model. (a): The architecture of the end-to-end model,
(b): The LSTM memory block in Bi-LSTM encoding layer, (c): The LSTM memory block in LSTM$_d$ decoding layer.
}
 \end{center}
\end{figure*}
In recent years, end-to-end model based on neural network is been widely used in sequence tagging task.
In this paper, we investigate an end-to-end model to produce the tags sequence as Figure \ref{model} shows.
It contains a bi-directional Long Short Term Memory (Bi-LSTM) layer to encode the input sentence
and a LSTM-based decoding layer with biased loss.
The biased loss can enhance the relevance of entity tags.

\noindent\textbf{The Bi-LSTM Encoding Layer.}
In sequence tagging problems, the Bi-LSTM encoding layer has been shown the effectiveness to
capture the semantic information of each word.
It contains forward lstm layer, backward lstm layer
and the concatenate layer.
The word embedding layer converts the word with 1-hot representation to an embedding vector.
Hence, a sequence of words can be represented as \begin{math}W = \{w_1,...w_{t},w_{t+1}...w_n\}\end{math},
where \begin{math}w_t \in\mathbb{R}^{d}\end{math} is the \emph{d}-dimensional word vector corresponding
to the \emph{t}-th word in the sentence and $n$ is the length of the given sentence.
After word embedding layer, there are two parallel LSTM layers: forward LSTM layer and backward LSTM layer.
The LSTM architecture consists of a set of recurrently connected subnets, known
as memory blocks. Each time-step is a LSTM memory block.
The LSTM memory block in Bi-LSTM encoding layer is used to compute current hidden vector $h_{t}$
based on the previous hidden vector $h_{t-1}$, the previous cell vector $c_{t-1}$ and the current input word embedding $w_t$.
Its structure diagram is shown in Figure \ref{model} $(b)$, and detail operations are defined as follows:
\begin{equation}\label{it}i_t = \delta (W_{wi}w_t+W_{hi}h_{t-1}+W_{ci}c_{t-1}+b_i),\end{equation}
\vspace{-0.8cm}
\begin{equation}\label{ft}f_t = \delta (W_{wf}w_t+W_{hf}h_{t-1}+W_{cf}c_{t-1}+b_f),\end{equation}
\vspace{-0.5cm}
\begin{equation}\label{zt}z_t = tanh (W_{wc}w_t+W_{hc}h_{t-1}+b_c),\end{equation}
\vspace{-0.5cm}
\begin{equation}\label{ct}c_t =f_tc_{t-1}+ i_t z_t,\end{equation}
\vspace{-0.8cm}
\begin{equation}\label{ot}o_t = \delta (W_{wo}w_t+W_{ho}h_{t-1}+W_{co}c_{t}+b_o),\end{equation}
\vspace{-0.5cm}
\begin{equation}\label{ht}h_t = o_t tanh(c_t),\end{equation}
where $i$, $f$ and $o$ are the input gate, forget gate and output
gate respectively, b is the bias term, c is the cell memory, 
and $W_{(.)}$  are the parameters.
For each word $w_t$, the forward LSTM layer will encode $w_t$
by considering the contextual information from word $w_1$ to $w_t$, which is marked as $\overrightarrow{h_t}$.
In the similar way, the backward LSTM layer will encode $w_t$ based on the contextual information from $w_n$ to $w_t$,
which is marked as $\overleftarrow{h_t}$.
Finally, we concatenate $\overleftarrow{h_t}$ and $\overrightarrow{h_t}$ to represent word $t$'s encoding information, denoted as ${h_t}= {[\overrightarrow{h_t},\overleftarrow{h_t}]}$.

\noindent\textbf{The LSTM Decoding Layer.} We also adopt a LSTM structure to produce the tag sequence.
When detecting the tag of word $w_t$, the inputs of decoding layer are:
${h_t}$ obtained from Bi-LSTM encoding layer, former predicted tag embedding ${T_{t-1}}$,
former cell value ${c^{(2)}_{t-1}}$,
and the former hidden vector in decoding layer ${h^{(2)}_{t-1}}$.
The structure diagram of the memory block in LSTM$_d$ is shown in Figure \ref{model} $(c)$,
and detail operations are defined as follows:
\begin{equation}\label{it}i^{(2)}_t = \delta (W^{(2)}_{wi}h_t+W^{(2)}_{hi}h^{(2)}_{t-1}+W_{ti}T_{t-1}+b^{(2)}_i),\end{equation}
\vspace{-0.8cm}
\begin{equation}\label{ft}f^{(2)}_t = \delta (W^{(2)}_{wf}h_t+W^{(2)}_{hf}h^{(2)}_{t-1}+W_{tf}T_{t-1}+b^{(2)}_f),\end{equation}
\vspace{-0.8cm}
\begin{equation}\label{zt}z^{(2)}_t = tanh (W^{(2)}_{wc}h_t+W^{(2)}_{hc}h^{(2)}_{t-1}+W_{tc}T_{t-1}+b^{(2)}_c),\end{equation}
\vspace{-0.4cm}
\begin{equation}\label{ct}c^{(2)}_t =f^{(2)}_tc^{(2)}_{t-1}+ i^{(2)}_t z^{(2)}_t,\end{equation}
\vspace{-0.8cm}
\begin{equation}\label{ot}o^{(2)}_t = \delta (W^{(2)}_{wo}h_t+W^{(2)}_{ho}h^{(2)}_{t-1}+W^{(2)}_{co}c_{t}^{(2)}+b^{(2)}_o),\end{equation}
\vspace{-0.5cm}
\begin{equation}\label{ht}h^{(2)}_t = o^{(2)}_t tanh(c_t^{(2)}),\end{equation}
\vspace{-0.5cm}
\begin{equation}{T_t} = W_{ts}h^{(2)}_t+b_{ts}.\end{equation}
The final softmax layer computes normalized entity tag probabilities based on the tag predicted vector ${T_{t}}$:
\begin{equation}\label{soft}y_t =  W_{y}T_t+b_y,\end{equation}
\vspace{-0.3cm}
\begin{equation}\label{tagp}
p_t^i=\frac{exp(y_t^{i})}{\sum\limits_{j=1}^{N_t}{exp(y_t^{j})}},
\end{equation}
where \begin{math}W_y\end{math} is the softmax matrix, $N_t$ is the total number of tags.
Because $T$ is similar to tag embedding and LSTM is capable of learning long-term dependencies,
the decoding manner can model tag interactions.

\noindent\textbf{The Bias Objective Function.}
We train our model to maximize the log-likelihood of the data and
the optimization method we used is RMSprop proposed by Hinton in \cite{hiton}.
The objective function can be defined as:
\begin{align*}L =& max \sum\limits_{j=1}^{|\mathbb D|} \sum\limits_{t=1}^{L_j}{(log(p_t^{(j)} = y_t^{(j)}|x_j,\Theta )\cdot I(O)} \\&{+
\alpha\cdot log(p_t^{(j)} = y_t^{(j)}|x_j,\Theta )\cdot (1-I(O)))},\end{align*}
where \begin{math}|\mathbb D|\end{math} is the size of training set,
\begin{math}L_j\end{math} is the length of sentence $x_j$, $y_t^{(j)}$ is the label of word $t$ in sentence $x_j$
and $p_t^{(j)}$ is the normalized probabilities of tags which defined in Formula \ref{tagp}.
Besides, $I(O)$ is a switching function to distinguish the loss of tag 'O' and relational tags that can indicate the results.
It is defined as follows:
$$ I(O)=\left\{
\begin{aligned}
1, & ~~if ~~~tag~=~'O' \\
0, & ~~if ~~~tag~\ne~'O'
.\end{aligned}
\right.
$$
$\alpha$ is the bias weight. The larger $\alpha$ is, the greater influence of relational tags on the model.

\section{Experiments}
\subsection{Experimental setting}
\textbf{Dataset}
To evaluate the performance of our methods, we use the public dataset NYT
\footnote{The dataset can be downloaded at: https://github.com/shanzhenren/CoType.
There are three data sets in the public resource and we only use the NYT dataset.
Because more than 50\% of the data in BioInfer has overlapping relations which is beyond the scope of this paper.
As for dataset Wiki-KBP, the number of relation type in the test set is more than that of the train set, which is also not suitable for a supervised training method.
Details of the data can be found in Ren's\cite{ren2016cotype} paper.
}
which is produced by distant supervision method \cite{ren2016cotype}.
A large amount of training data can be obtained by means of distant supervision methods without manually labeling.
While the test set is manually labeled to ensure its quality.
In total, the training data contains $353k$ triplets,
and the test set contains $3,880$ triplets.
Besides, the size of relation set is $24$.

\noindent\textbf{Evaluation}
We adopt standard Precision (Prec), Recall (Rec) and F1 score to evaluate the results.
Different from classical methods,
our method can extract triplets without knowing the information of entity types.
In other words, we did not use the label of entity types to train the model,
therefore we do not need to consider the entity types in the evaluation.
A triplet is regarded as correct when its relation type and the head offsets of two corresponding entities are both correct.
Besides, the ground-truth relation mentions are given and ``None'' label is excluded as \cite{ren2016cotype,li2014,miwa2016end} did.
We create a validation set by randomly sampling $10\%$ data from test set and use the remaining data as evaluation based on  \cite{ren2016cotype}'s suggestion. We run 10 times for each experiment then report the average results and their
standard deviation
as Table \ref{result} shows.

\noindent\textbf{Hyperparameters}
Our model consists of a Bi-LSTM encoding layer and a LSTM decoding layer with bias objective function.
The word embeddings used in the encoding part are initialed by running word2vec\footnote{https://code.google.com/archive/p/word2vec/} \cite{w2v} on NYT training corpus.
The dimension of the word embeddings is $d=300$. We regularize our network using
dropout on embedding layer and the dropout ratio is $0.5$.
The number of lstm units in encoding layer is $300$ and the number in decoding layer is $600$.
The bias parameter $\alpha$ corresponding to the results in Table \ref{result} is $10$.
\begin{table*}[ht]
\centering
\begin{tabular}{cccc}\hlinewd{1.2pt}
Methods	& ~~\emph{Prec.}~~&~~\emph{Rec.}~~ &~~\emph{F1}~~	\\\hlinewd{1.2pt}
FCM	&0.553	&0.154	&0.240\\
DS+logistic&	0.258	&0.393	&0.311\\
LINE	&0.335	&0.329	&0.332\\\hline
MultiR	&0.338	&0.327	&0.333\\
DS-Joint	&0.574	&0.256	&0.354\\
CoType	&0.423	&\textbf{0.511	}&0.463\\\hline
LSTM-CRF	&$\textbf{0.693} \pm \textbf{0.008}	$&$0.310 \pm 0.007	$&$0.428 \pm 0.008$\\
LSTM-LSTM	&$0.682 \pm 0.007	$&$0.320 \pm 0.006	$&$0.436 \pm 0.006$\\
\textbf{LSTM-LSTM-Bias	}&$0.615 \pm 0.008	$&$0.414\pm 0.005	$&$\textbf{0.495} \pm \textbf{0.006}$\\\hlinewd{1.2pt}
\end{tabular}
\caption{\label{result}The predicted results of different methods on extracting both entities and their relations.
The first part (from row $1$ to row $3$) is the pipelined methods and the second part (row $4$ to $6$) is the jointly extracting methods.
Our tagging methods are shown in part three (row $7$ to $9$). In this part, we not only report the results of precision, recall and F1, we also
compute their standard deviation.
  }
\end{table*}

\noindent\textbf{Baselines}
We compare our method with several classical triplet extraction methods,
which can be divided into the following categories: the pipelined methods, the jointly extracting methods and the end-to-end methods based our tagging scheme.

For the pipelined methods, we follow \cite{ren2016cotype}'s settings:
The NER results are obtained by CoType \cite{ren2016cotype} then several classical relation classification methods
 are applied to detect the relations.
These methods are:
(1) DS-logistic \cite{dslogistic} is a distant supervised and feature based method, which combines the advantages of supervised IE and unsupervised IE features;
 (2)  LINE \cite{line} is a network embedding method, which is suitable for arbitrary types of information networks;
(3) FCM \cite{fcm} is a compositional model that combines lexicalized linguistic context and word embeddings for relation extraction.

The jointly extracting methods used in this paper are listed as follows:
(4) DS-Joint \cite{li2014} is a supervised method, which jointly extracts entities and relations using structured perceptron on human-annotated dataset;
(5) MultiR \cite{mutilr} is a typical distant supervised method based on multi-instance learning algorithms to combat the noisy training data;
(6) CoType \cite{ren2016cotype} is a domain independent framework by jointly embedding entity mentions, relation mentions, text features and type labels into meaningful representations.

In addition, we also compare our method with two classical end-to-end tagging models:
LSTM-CRF \cite{lstm-crf} and LSTM-LSTM \cite{supertagging}.
LSTM-CRF is proposed for entity recognition by using a bidirectional LSTM
to encode input sentence and a conditional random fields to predict the entity tag sequence.
Different from LSTM-CRF, LSTM-LSTM uses a LSTM layer to decode the tag sequence instead of CRF.
They are used for the first time to jointly extract entities and relations based on our tagging scheme.

\subsection{Experimental Results}
We report the results of different methods as shown in Table \ref{result}.
It can be seen that our method, LSTM-LSTM-Bias, outperforms all other methods in F1 score
and achieves a $3\%$ improvement in $F1$ over the best method CoType \cite{ren2016cotype}.
It shows the  effectiveness of our proposed method.
Furthermore, from Table \ref{result}, we also can see that the jointly extracting methods are better than pipelined methods,
and the tagging methods are better than most of the jointly extracting methods.
It  also validates the validity of our tagging scheme for the task of jointly extracting entities and relations.

\begin{table*}[ht] 
\centering
\begin{tabular}{c|ccc|ccc|ccc}\hlinewd{1.2pt}
Elements          &\multicolumn{3}{|c|}{E1} &\multicolumn{3}{|c|}{E2} &\multicolumn{3}{c}{(E1,E2)} \\\cline{1-10}
PRF
& \emph{Prec.}&\emph{Rec.} &\emph{F1}
& \emph{Prec.}&\emph{Rec.} &\emph{F1}
& \emph{Prec.}&\emph{Rec.} &\emph{F1} \\\hlinewd{1.2pt}
LSTM-CRF	&\textbf{0.596}	&0.325	&0.420&	0.605&	0.325&	0.423&\textbf{0.724}	&0.341 	&0.465 \\
LSTM-LSTM	&0.593&	0.342&	0.434&	\textbf{0.619}&	0.334&	0.434 &	0.705 &0.340 &	0.458\\
LSTM-LSTM-Bias&	0.590&\textbf{0.479}&	\textbf{0.529}&	0.597&	\textbf{0.451}&  \textbf{0.514}&0.645 &\textbf{0.437}&\textbf{0.520} \\
\hlinewd{1.2pt}
\end{tabular}
\caption{\label{ana1}The predicted results of triplet's elements based on our tagging scheme.
  }
\end{table*}
When compared with the traditional methods,
the precisions of the end-to-end models are significantly improved.
But only LSTM-LSTM-Bias can be better to balance the precision and recall. The reason may be that
these end-to-end models all use a Bi-LSTM encoding input sentence and different neural networks to decode the results.
The methods based on neural networks can well fit the data.
Therefore, they can learn the common features of the training set well and may lead to the lower expansibility.
We also find that the LSTM-LSTM model is better than LSTM-CRF model based on our tagging scheme.
Because, LSTM is capable of learning long-term dependencies and
CRF \cite{crf} is good at capturing the  joint probability of the entire sequence of labels.
The related tags may have a long distance from each other.
Hence, LSTM decoding manner is a little  better than CRF.
LSTM-LSTM-Bias adds a bias weight to enhance the effect of entity tags and weaken the effect of invalid tag.
Therefore, in this tagging scheme, our method can be better than the common LSTM-decoding methods.

%
\section{Analysis and Discussion}
\subsection{Error Analysis}
In this paper, we focus on extracting triplets composed of two entities and a relation.
Table \ref{result} has shown the predict results of the task. It treats an triplet is correct only when the relation type and
the head offsets of two corresponding entities are both correct.
In order to find out the factors that affect the results of end-to-end models,
we analyze the performance on predicting each element in the triplet as Table \ref{ana1} shows.
$E1$ and $E2$ represent the performance on predicting each entity, respectively.
If the head offset of the first entity is correct, then the instance of $E1$ is correct, the same to $E2$.
Regardless of relation type, if the head offsets of two corresponding entities are both correct,  the instance of $(E1,E2)$ is correct.

As shown in Table \ref{ana1}, $(E1,E2)$ has higher precision when compared with $E1$ and $E2$.
But its recall result is lower than $E1$ and $E2$.
It means that some of the predicted entities do not form a pair.
They only obtain $E1$ and do not find its corresponding $E2$,
or obtain $E2$ and do not find its corresponding $E1$.
Thus it leads to the prediction of more single $E$ and less $(E1,E2)$ pairs.
Therefore, entity pair $(E1,E2)$ has higher precision and lower recall than single $E$.
Besides, the predicted results of $(E1,E2)$ in Table \ref{ana1} have about 3\% improvement when compared predicted results in Table \ref{result}, which means that 3\% of the test data is predicted to be wrong
because the relation type is predicted to be wrong.
\subsection{Analysis of Biased Loss}
\begin{figure}[h]
 \begin{center}
\includegraphics[width=7cm]{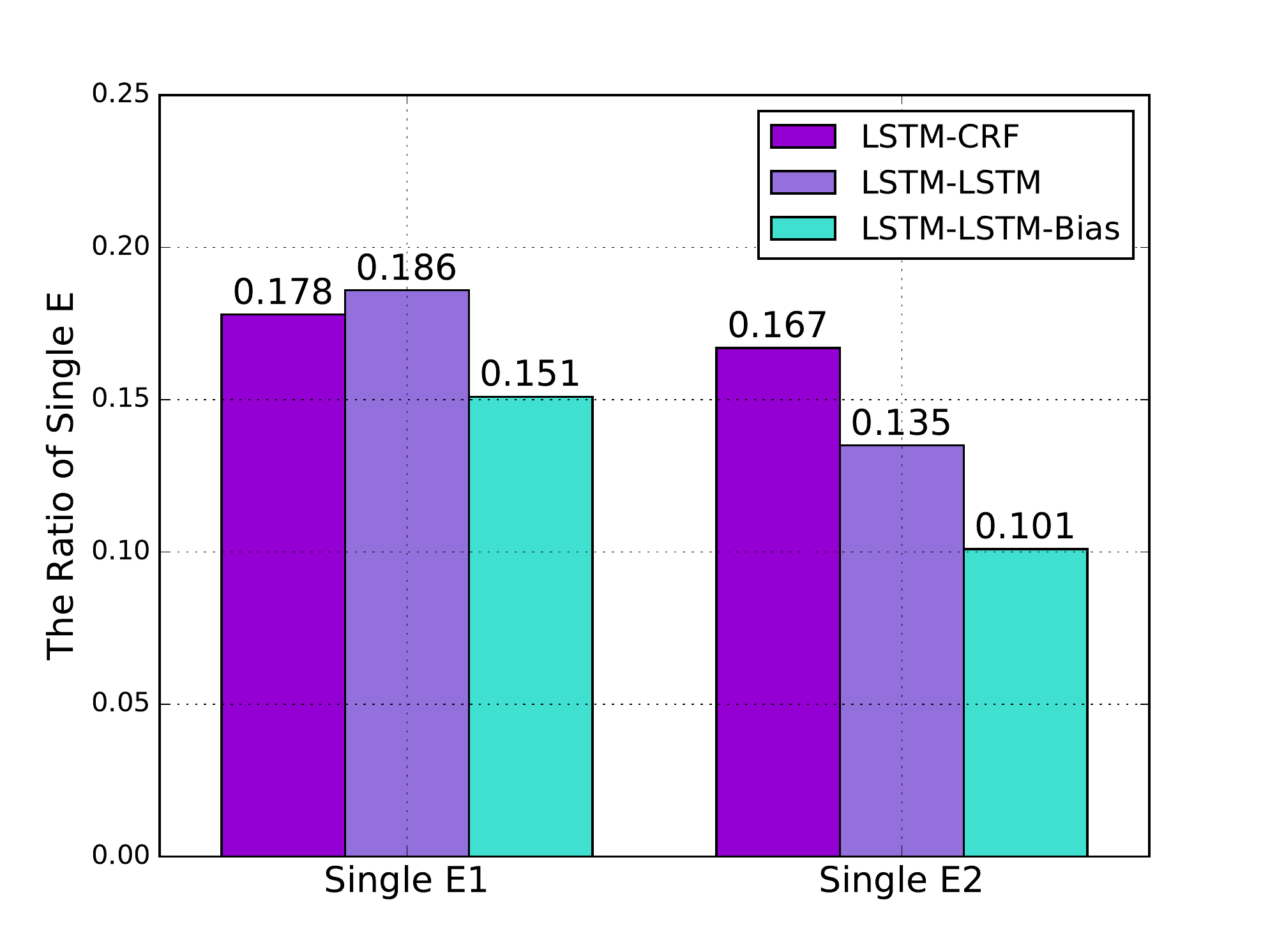}
\caption{\label{singlee}The ratio of predicted single entities for each method.
The higher of the ratio the more entities are left.}
 \end{center}
\end{figure}

Different from LSTM-CRF and LSTM-LSTM,
our approach is biased towards relational labels to enhance links between entities.
In order to further analyze the effect of the bias objective function,
we visualize the ratio of predicted single entities for each end-to-end method as Figure \ref{singlee}.
The single entities refer to those who cannot find their corresponding entities.
Figure \ref{singlee} shows whether it is E1 or E2, our method can get a relatively low ratio on the single entities.
It means that our method can effectively associate two entities when compared LSTM-CRF and LSTM-LSTM which pay little attention to
the relational tags.

 \begin{figure}[h]
 \begin{center}
\includegraphics[width=7cm]{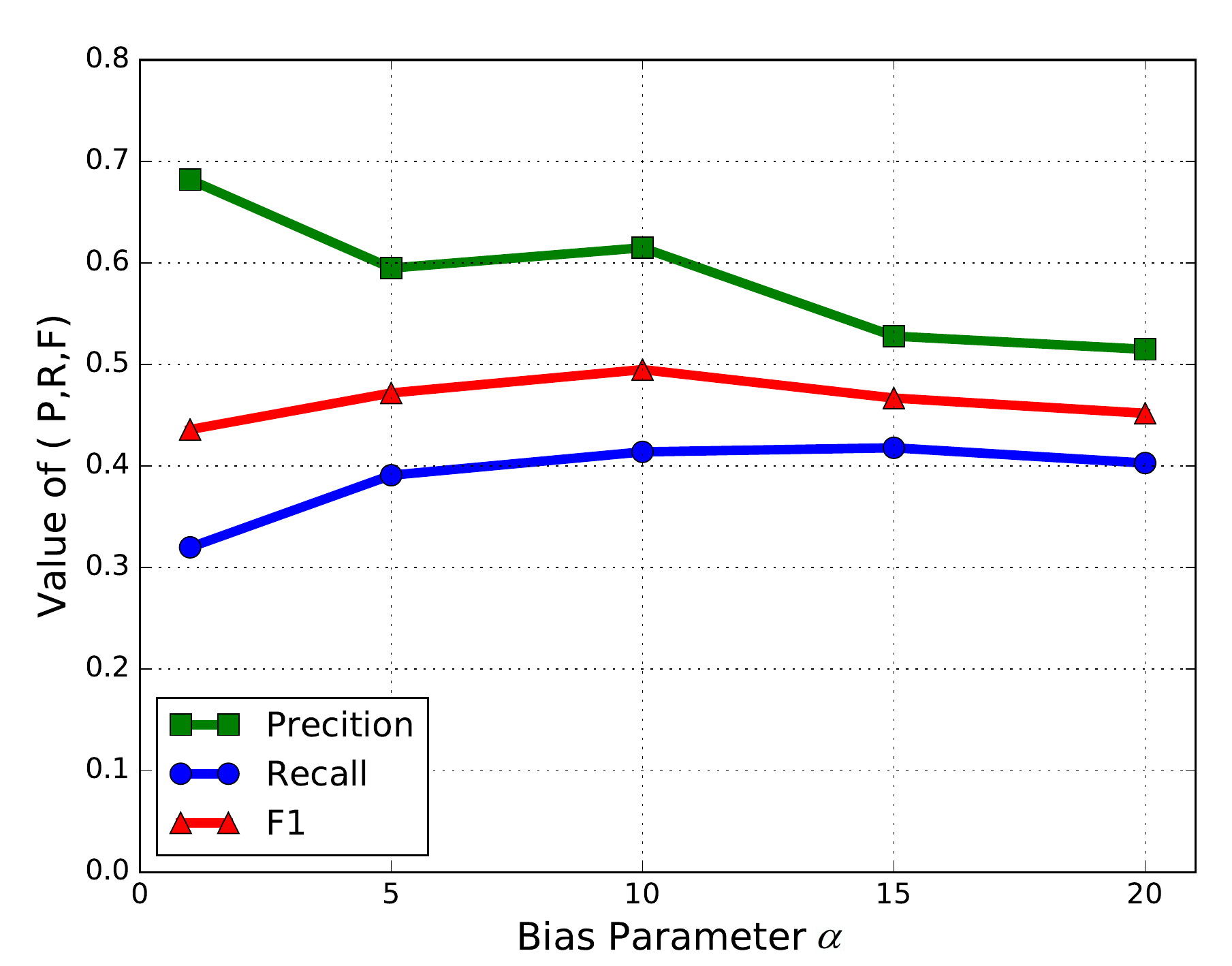}
\caption{\label{loss}The results predicted by LSTM-LSTM-Bias on different bias parameter $\alpha$.}
 \end{center}
\end{figure}

Besides, we also change the Bias Parameter $\alpha$ from $1$ to $20$, and the predicted results are shown in Figure \ref{loss}.
 If $\alpha$ is too large, it will affect the accuracy of prediction and if $\alpha$ is too small, the recall will decline.
 When $\alpha = 10$, LSTM-LSTM-Bias can balance the precision and recall, and can achieve the best F1 scores.
\subsection{Case Study}
\begin{table*}[ht]
\centering
\begin{tabular}{cl}\hlinewd{1.2pt}\hline
Standard S1&	\tabincell{l}{
\textcolor{blue}{\textbf{[Panama City Beach]$_{E2contain}$}} has  condos , but the area was one of only two\\
in \textcolor{blue}{{\textbf{[Florida]$_{E1contain}$}}} where sales rose in March , compared with a year earlier.}\\\hline
LSTM-LSTM&\tabincell{l}{
\textcolor{red}{Panama City Beach} has  condos , but the area was one of only two
in  \\ \textcolor{blue}{{\textbf{[Florida]$_{E1contain}$}}} where sales rose in March , compared with a year earlier.}\\\hline
LSTM-LSTM-Bias	&\tabincell{l}{
\textcolor{blue}{\textbf{[Panama City Beach]$_{E2contain}$}} has  condos , but the area was one of only two\\
in \textcolor{blue}{{\textbf{[Florida]$_{E1contain}$}}} where sales rose in March , compared with a year earlier.}\\\hline \hline
Standard S2&	\tabincell{l}{
All came from \textcolor{blue}{\textbf{[Nuremberg]$_{E2contain}$}} , \textcolor{blue}{\textbf{[Germany]$_{E1contain}$}} , a center of brass \\production since the Middle Ages.}\\\hline
LSTM-LSTM&\tabincell{l}{
All came from Nuremberg , \textcolor{blue}{\textbf{[Germany]$_{E1contain}$}} , a center of brass production\\ since the \textcolor{red}{\textbf{[Middle Ages]$_{E2contain}$}}.}\\\hline
LSTM-LSTM-Bias	&\tabincell{l}{
All came from Nuremberg , \textcolor{blue}{\textbf{[Germany]$_{E1contain}$}} , a center of brass production\\ since the \textcolor{red}{\textbf{[Middle Ages]$_{E2contain}$}}.}\\\hline \hline
Standard S3&	\tabincell{l}{
\textcolor{blue}{\textbf{[Stephen A.]$_{E2CF}$}} , the co-founder of the \textcolor{blue}{\textbf{[Blackstone Group]$_{E1CF}$}},  which\\ is in the process of going public , made \$ 400 million last year.}\\\hline
LSTM-LSTM&\tabincell{l}{
\textcolor{red}{\textbf{[Stephen A.]$_{E1CF}$}} , the co-founder of the \textcolor{blue}{\textbf{[Blackstone Group]$_{E1CF}$}}, which\\  is in the process of going public , made \$ 400 million last year.}\\\hline
LSTM-LSTM-Bias&\tabincell{l}{
\textcolor{red}{\textbf{[Stephen A.]$_{E1CF}$}} , the co-founder of the \textcolor{red}{\textbf{[Blackstone Group]$_{E2CF}$}},  which\\ is in the process of going public , made \$ 400 million last year.}\\\hline \hline
\hlinewd{1.2pt}
\end{tabular}
\caption{\label{case}Output from different models. Standard $S_i$ represents the gold standard of sentence $i$.
The blue part is the correct result, and the red one is the wrong one. $E1CF$ in case '$3$' is short for $E1_{Company-Founder}$.
  }
\end{table*}
In this section, we observe the prediction results of end-to-end methods, and then select several representative examples to illustrate the advantages and disadvantages of the methods as Table \ref{case} shows.
Each example contains three row, the first row is the gold standard, the second and the third rows are the extracted results of model LSTM-LSTM and LSTM-LSTM-Bias respectively.

$S1$ represents the situation that the distance between the two interrelated entities is far away from each other,
which is more difficult to detect their relationships. When compared with LSTM-LSTM,
LSTM-LSTM-Bias uses a bias objective function which enhance the relevance between entities.
Therefore, in this example, LSTM-LSTM-Bias can extract two related entities, while LSTM-LSTM can only extract one entity of ``Florida'' and can not detect entity ``Panama City Beach''.

$S2$ is a negative example that shows these methods may mistakenly predict one of the entity.
There are no indicative words between entities $Nuremberg$ and $Germany$.
Besides, the patten ``a * of *'' between $Germany$ and $Middle Ages$
may be easy to mislead the models that there exists a relation of ``Contains'' between them.
The problem can be solved by adding some samples of this kind of expression patterns to the training data.

$S3$ is a case that models can predict the entities' head offset right, but the relational role is wrong.
LSTM-LSTM treats both ``Stephen A. Schwarzman'' and ``Blackstone Group'' as entity $E1$,
and
can not find its corresponding $E2$.
Although, LSTM-LSMT--Bias can find the entities pair $(E1,E2)$, it reverses the roles of ``Stephen A. Schwarzman'' and ``Blackstone Group''.
It shows that LSTM-LSTM-Bias is able to better on predicting entities pair,
but it remains to be improved in distinguishing the relationship between the two entities.
\section{Conclusion}
In this paper, we propose a novel tagging scheme and investigate the end-to-end models to jointly extract entities and relations.
The experimental results show the  effectiveness of our proposed method.
But it still has shortcoming on the identification of the overlapping relations.
In the future work, we will replace the softmax function in the output layer
with multiple classifier,
so that a word can has multiple tags.
In this way, a word can appear in multiple triplet results, which can solve the problem of overlapping relations.
Although, our model can enhance the effect of entity tags,
the association between two corresponding entities still requires refinement in next works.
\section*{Acknowledgments}
We thank Xiang Ren for dataset details and helpful discussions.
This work is also supported by the National
High Technology Research and Development Program
of China (863 Program) (Grant No. 2015AA015402), the National Natural Science Foundation of China (No. 61602479) and
the NSFC project 61501463.
\bibliography{acl2017}
\bibliographystyle{acl_natbib}
\appendix

\end{document}